\documentclass[letterpaper, 10 pt, conference]{ieeeconf}  % Comment this line out if you need a4paper

\IEEEoverridecommandlockouts                              % This command is only needed if 
\usepackage{graphics} % for pdf, bitmapped graphics files
\usepackage{epsfig} % for postscript graphics files
\usepackage{mathptmx} % assumes new font selection scheme installed
\usepackage{times} % assumes new font selection scheme installed
\usepackage{amsmath} % assumes amsmath package installed
\usepackage{amssymb}  % assumes amsmath package installed
\usepackage{kotex}
\usepackage{tabularx}
\usepackage{booktabs}
\usepackage{graphicx}
\usepackage{algorithm}
\usepackage{algpseudocode}
\usepackage{subcaption}
\usepackage[font=footnotesize,labelfont=bf]{caption}
\usepackage{placeins}
\usepackage{afterpage}
\usepackage{multirow}
\usepackage{comment}
\usepackage{cite}
\usepackage{xcolor}

\title{\LARGE \bf
Multitask Learning for Multiple Recognition Tasks: A Framework for Lower-limb Exoskeleton Robot Applications
}

\author{Joonhyun Kim$^{1}$, Seongmin Ha$^{2}$, Dongbin Shin$^{3}$, Seoyeon Ham$^{4}$ Jaepil Jang$^{4}$, and Wansoo Kim$^{5,*}$% <-this % stops a space
\thanks{This work was supported by Institute for Information \& communications Technology Promotion(IITP) grant funded by the Korea government(MSIP) (No.2022-0-00860,Development of Solution Technology for Personalized Gait Control and Performance Evaluation of Lower-Limb Robotic Exoskeleton through Artificial Intelligence/Big Data)}% <-this % stops a space
\thanks{$^{1}$J. Kim is with Department of Applied Artificial Intelligence, Hanyang University 55, Hanyangdaehak-ro, Sangnok-gu, Ansan-si, Gyeonggi-do, Republic of Korea ({\tt\small e-mail:ralwnsgus4715@gmail.com}).}%
\thanks{$^{2}$S. Ha is with Department of Mechatronics Engineering, Hanyang University, Republic of Korea.}%
\thanks{$^{3}$D. Shin is with Hexar Humancare Co., Ltd., Republic of Korea.}%
\thanks{$^{4}$S. Ham and J. Jang are with Department of Interdisciplinary Robot Engineering Systems, Hanyang University, Republic of Korea.}
\thanks{$^{5*}$W. Kim is with Robotics Department, Hanyang University ERICA, Ansan-si, Gyeonggi-do, Republic of Korea ({\tt\small e-mail:wansookim@hanyang.ac.kr}).}
\thanks{$^{*}$Corresponding Author} %
}

\begin{document}

\maketitle
\thispagestyle{empty}
\pagestyle{empty}

%%%%%%%%%%%%%%%%%%%%%%%%%%%%%%%%%%%%%%%%%%%%%%%%%%%%%%%%%%%%%%%%%%%%%%%%%%%%%%%%
\begin{abstract}
To control the lower-limb exoskeleton robot effectively, it is essential to accurately recognize user status and environmental conditions. Previous studies have typically addressed these recognition challenges through independent models for each task, resulting in an inefficient model development process. In this study, we propose a Multitask learning approach that can address multiple recognition challenges simultaneously. This approach can enhance data efficiency by enabling knowledge sharing between each recognition model. We demonstrate the effectiveness of this approach using Gait phase recognition (GPR) and Terrain classification (TC) as examples, the most conventional recognition tasks in lower-limb exoskeleton robots. We first created a high-performing GPR model that achieved a Root mean square error (RMSE) value of 2.345 \(\pm\) 0.08\% and then utilized its knowledge-sharing backbone feature network to learn a TC model with an extremely limited dataset. Using a limited dataset for the TC model allows us to validate the data efficiency of our proposed Multitask learning approach. We compared the accuracy of the proposed TC model against other TC baseline models. The proposed model achieved 99.5 \(\pm\) 0.044\% accuracy with a limited dataset, outperforming other baseline models, demonstrating its effectiveness in terms of data efficiency. Future research will focus on extending the Multitask learning framework to encompass additional recognition tasks.
\end{abstract}

%%%%%%%%%%%%%%%%%%%%%%%%%%%%%%%%%%%%%%%%%%%%%%%%%%%%%%%%%%%%%%%%%%%%%%%%%%%%%%%%
\section{Introduction}
To control the robot effectively, appropriate interaction with the environment is necessary, requiring accurate recognition of environmental changes. This principle also applies to lower-limb exoskeleton robots, which are wearable robotic devices designed to provide support, assistance, and augmentation to human legs, enhancing mobility and strength\cite{trend1,hri_prof_kim}.

From the perspective of lower-limb exoskeleton robots, not only the physical world but also the human element is considered a part of the external environment, which constantly applies force while the robot must assist in directing movement. Since lower-limb exoskeleton robots are always attached to humans, effectively addressing recognition problems becomes essential.

\begin{figure}[!t]
\centering
\includegraphics[width=1.0\linewidth]{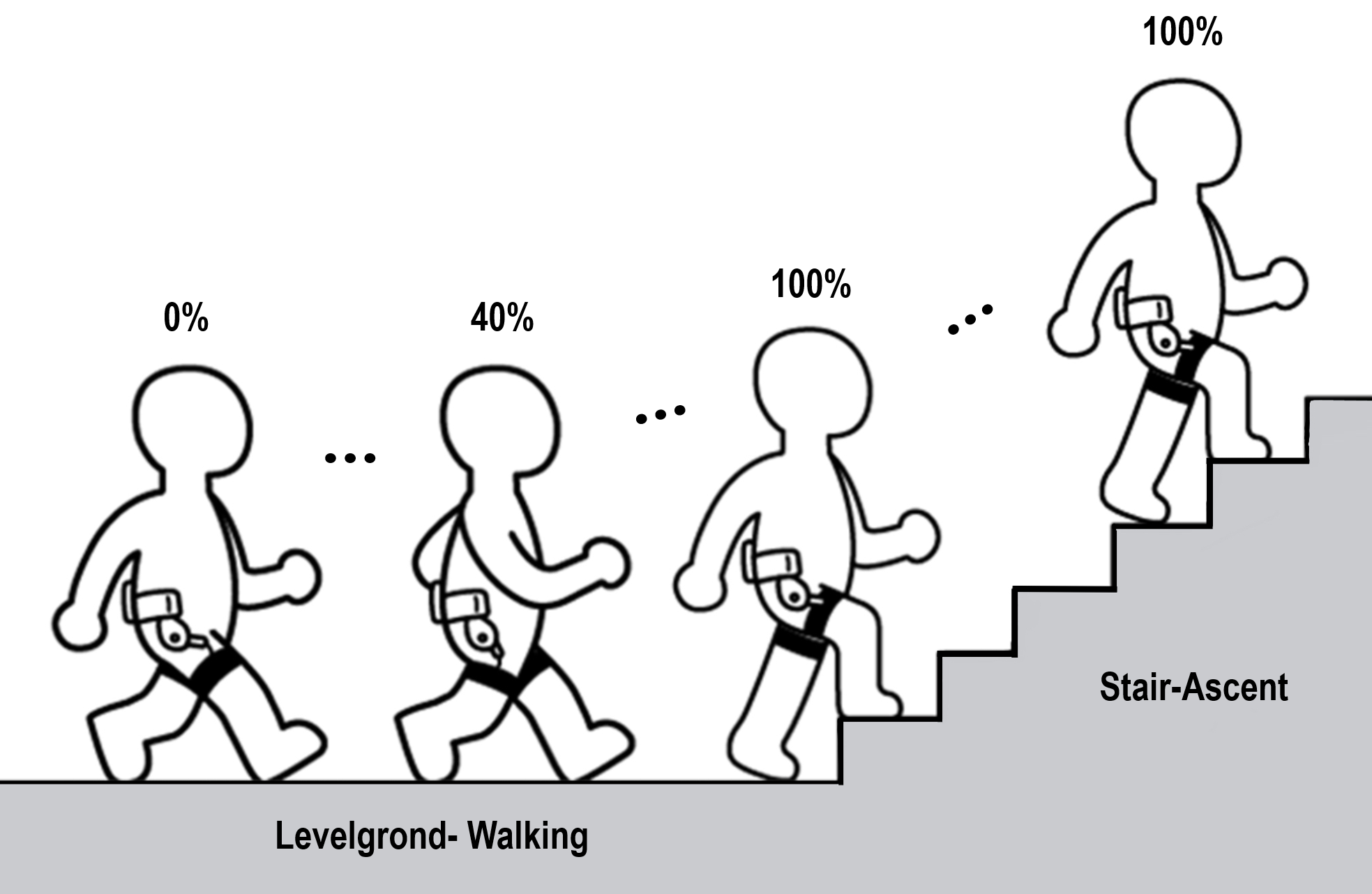}
\caption{TC\& GPR, the most conventional recognition problems of lower-limb exoskeleton robot control situation. TC refers to the identification of the terrain on which the robot's user is moving, while GPR means recognizing the user's gait phase between 0\% and 100\%.}
\label{fig:TR_GPR_exampleimg}
\end{figure}

This paper focuses on a more effective approach to handling multiple recognition problems for lower-limb exoskeleton robots. Until now, models solving these recognition problems have been developed separately to work effectively in each task\cite{machinelearning_gpr1,machinelearning_tc1}. However, this presents challenges in configuring new model structures and data processing algorithms suitable for each model, which can be time-consuming and inefficient. Furthermore, collecting large data samples related to human motion is challenging due to factors such as organizing separate experimental protocols, equipment battery longevity, and subject fatigue \cite{Georgiadata}. Therefore, achieving high data efficiency is crucial when developing machine learning models for exoskeleton robots.

To achieve high data efficiency, this paper proposes a Multitask learning technique that can address multiple recognition tasks simultaneously. Multitask learning, which originates from the Representation learning paradigm, is a machine learning approach that enables a model to efficiently learn and perform related tasks even with small data samples by utilizing the knowledge for a particular task to learn another related task \cite{representation_review}.

In this study, we propose a Multitask learning framework to demonstrate two conventional recognition problems for lower-limb exoskeleton robots: Gait phase recognition (GPR) and Terrain classification (TC) as illustrated in Figure \ref{fig:TR_GPR_exampleimg} \cite{multisensorgaitphase3,gaitphase_review,terrain_classification}.
GPR is the task of determining the specific phase of a walking cycle, which is essential for controlling lower-limb exoskeleton robots during various gait phases. GPR can be performed in either a discrete or continuous manner. Discrete GPR focuses on identifying distinct gait events, such as heel strike, mid-stance, heel off, and swing, while continuous GPR estimates the ongoing progression of the gait cycle, offering more detailed information for controlling the exoskeleton robot \cite{discrete, continuous}. On the other hand, TC is the task of identifying the type of surface on which the user is walking, such as stairs, ramps, or level ground, enabling the robot to adapt its assistance strategy differently to diverse environmental conditions \cite{TC}.

% Our hypothesis suggests that the TC task model can learn effectively even with smaller data samples through a well-trained GPR model by using the proposed Multitask learning framework. 
We hypothesize that the proposed Multitask learning framework can enhance the learning effectiveness of the TC task model, even with smaller data samples, by using a well-trained GPR model. 
% to demonstrate the effectiveness of our Multitask learning approach for recognition tasks in lower-limb exoskeletons, suggesting that the TC task model can learn effectively even with smaller data samples through a well-trained GPR model. 
In order to explore this hypothesis, we first develop a high-performing GPR Convolutional neural networks (CNN) model and then utilize some layers of its network to train other head networks that address a new task, TC \cite{cnn_survey}.

To validate our proposed framework, we compare the performance of our proposed TC model with that of other baseline models learned without a feature network. The comparison is conducted using a limited dataset to effectively validate data efficiency. As a result, the proposed model outperformed the other baseline models, demonstrating its potential to overcome data scarcity and tackle additional recognition challenges in lower-limb exoskeleton robots.

The remainder of this paper is organized as follows: Section II introduces the background and motivation for our research, focusing on the concept of Multitask learning and its application in lower-limb exoskeleton robots. Section III presents the proposed Multitask model implementation, detailing the input pipelining algorithm, model structure, labeling process for the GPR model, and training procedures. Section IV covers the experiments \& results to evaluate the proposed approach. Finally, Sections V presents the conclusion, including a summary and the potential implications of our research, along with future research directions.

\section{Background \& Motivation}

\subsection{What is Multitask Learning?}
Traditional machine learning typically involves extracting features from raw data and making predictions based on these features \cite{feature_engineering_history}. In the past, most features were extracted using expert domain knowledge, which meant that developing a high-performance machine learning algorithm required an expert familiar with the problem to create an appropriate feature representation. This feature engineering process was labor-intensive and relied more on understanding the data than on the quantity of data, making it difficult to fully utilize big data.

However, recent advancements in Deep neural networks(DNN) have enabled the extraction of rich representation features that surpass human domain knowledge \cite{DeepLearning}. By utilizing big data, models can achieve high performance in feature extraction, allowing the model to extract the necessary characteristics for a given task without requiring high-level domain expertise.

This well-trained DNN model has another potential advantage: it can share a feature network between tasks. Feature networks of well-trained models can be beneficial for learning related tasks in terms of data efficiency and model performance. This process, known as Representation learning, involves applying the feature network to other related tasks \cite{representation_review}.

\begin{figure}[!t]
\centering
\includegraphics[width=1.0\linewidth]{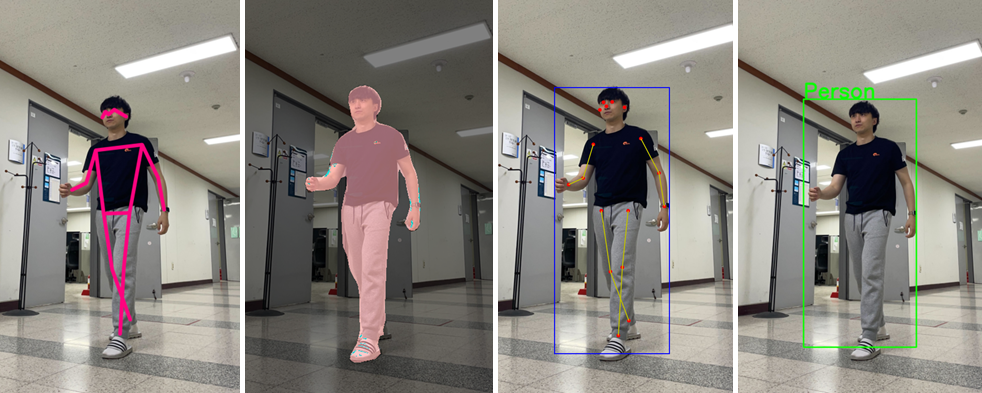}
\caption{In the field of computer vision, a shared feature representation network enables the efficient execution of various tasks. The examples show that multiple tasks being performed using the same mask R-CNN feature representation network. By leveraging the shared feature representation, tasks such as human keypoint estimation, image segmentation, bounding box generation, and image classification can be performed efficiently.}
\label{fig:rcnn_exampleimage}
\end{figure}

Multitask learning is one of the learning paradigms within the representation paradigm. It involves implementing a model that allows a shared feature network to perform multiple tasks simultaneously, resulting in benefits in terms of data efficiency.

The field of computer vision has been the most successful in applying Multitask learning by using CNNs. By making the convolutional kernel's weight a learnable parameter and updating it through backpropagation, CNNs can create rich and characteristic feature representations of data. Performing Multitask learning with these feature representations enables the simultaneous tackling of various computer vision tasks, such as creating bounding boxes, segmentation, key point estimation, and image captioning, more efficiently\cite{rcnn1, rcnn2, rcnn3, Multitask_review}, as described in Figure \ref{fig:rcnn_exampleimage}.

\subsection{Application to Lower-limb Exoskeleton Robots}
To effectively control lower-limb exoskeleton robots, environmental recognition problems must be addressed. Due to the nature of lower-limb exoskeleton robots, they are always attached to the wearer's body, and from the robot's perspective, humans are also part of the environment. Therefore, these recognition problems become even more crucial to consider. 

Collecting large amounts of data related to human motion can be challenging, leading to inefficiencies and additional resources when developing separate models for each recognition problem. Achieving high data efficiency is critical when developing machine learning models for exoskeleton robots. Therefore, we develop the Multitask learning framework that could be an attractive solution to the data efficiency issue \cite{pretraining_adv}. Multitask learning allows the model to utilize the already learned features and apply them to other related tasks, resulting in improved data efficiency and model performance.

By enabling information sharing between different tasks, we hypothesize that the cost required for learning new tasks could be reduced. We demonstrate this by using GPR and TC tasks as examples and proving the effectiveness of our Multitask learning approach, the details of implementation will be covered in the remainder of this paper.

\section{Multitask Learning Implementation}
\begin{figure}[!t]
\centering
    \begin{minipage}{1.0\linewidth}
        \centering
        \begin{subfigure}{1.0\linewidth}
            \centering
            \includegraphics[width=1.01\linewidth]{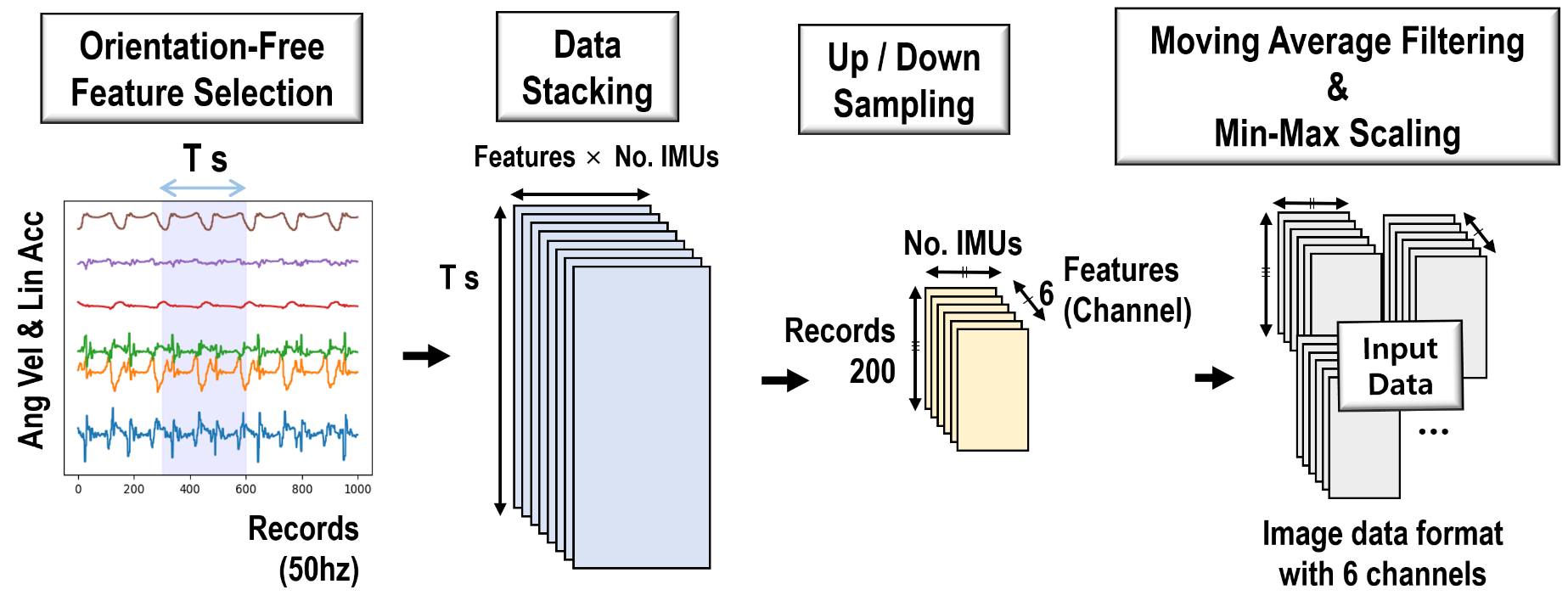}
            \caption{This figure outlines the input pipelining algorithm process. Orientation-free features, such as Linear acceleration(Lin Acc) and Angular velocity(Ang Vel), are selected and segmented over a duration of T seconds. And then, the stacked data is refined through Up/Down Sampling, moving average filtering, and min-max scaling. The format of the input data for the Multitask model, illustrated in Figure \ref{subfig:model_structure}, resembles a six-channel image.}
            \label{subfig:input_pipeline}
        \end{subfigure}
        \hfill
        \begin{subfigure}{1.0\linewidth}
            \centering
            \includegraphics[width=1.01\linewidth,height=0.5\linewidth]{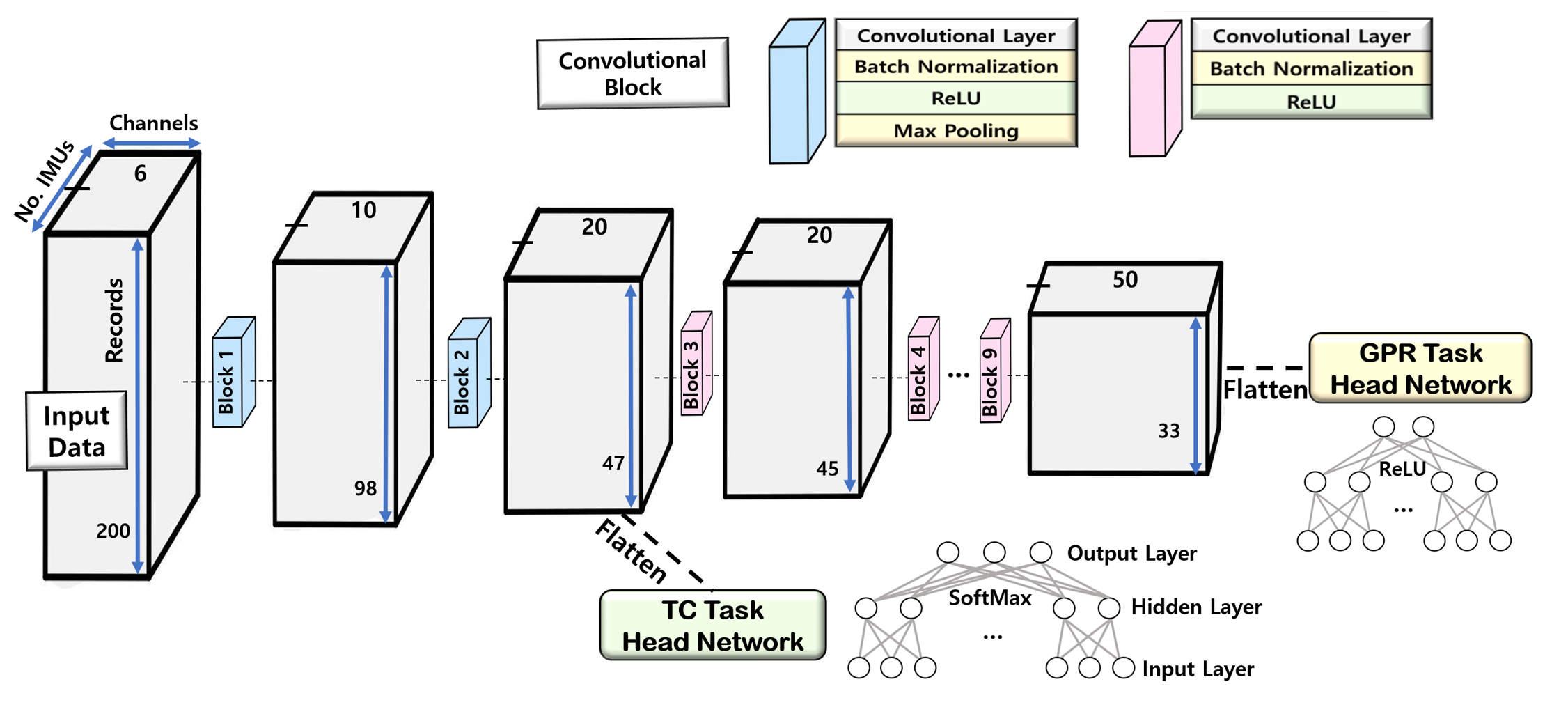}
            \caption{This illustration presents the overall model architecture and depicts the flow of input data as it passes through the model. The GPR head network connects to the 9th convolutional block of the feature network, while the TC head network links to the 2nd convolutional block of the same feature network, as detailed in Table \ref{table:featnet_gaitphase},\ref{table:featnet_terrain}.
            The head network consists of an MLP structure, consisting of two and three nodes in the Output Layer, respectively.}
            \label{subfig:model_structure}
        \end{subfigure}
    \end{minipage}
    \caption{Illustration of the Input pipelining algorithm and Model structure}
\end{figure}
\subsection{Input Pipelining Algorithm}
In order to implement the Multitask learning approach, we first develop an input pipelining algorithm that is suited for CNN models. The algorithm we propose in this paper has an input data format of batch size × (6 channels × 200 records × number of IMUs). This format can be viewed as a three-dimensional input value in the form of an image with six channels, which is expected to work well with convolutional kernels. 

Our research prioritizes user convenience by using only one IMU sensor attached to the left thigh, resulting in an input data format of batch size × (6 channels × 200 records × 1 IMU). The input pipelining algorithm consists of four steps, which are detailed below, and the overall process is illustrated in Figure \ref{subfig:input_pipeline}.

\subsubsection{Feature Selection}
Features are selected by choosing orientation-free features, linear acceleration, and angular velocity values along the $x, y, z$ axes in the IMU sensor's local reference frame. This ensures that the feature values are detected consistently, regardless of the user's standing direction.

\subsubsection{Data Stacking}
Data is stacked over a time window \textit{T} seconds and then segmented in a 2D grid-like format. The accumulated data over a set duration are then used as input data. The variable \textit{T} seconds can be used to augment the data, allowing for more flexibility in data representation and potentially enhancing model performance. The \textit{T} values for data augmentation are 1.5, 1.6, and 1.7.

\subsubsection{Up/Down Sampling}
Up/Down sampling is performed to maintain a constant input data size of 200 records, ensuring consistency across different input samples.

\subsubsection{Smoothing \& Normalization}
To smooth the data and prevent discrete points from negatively impacting learning, an moving average filter is implemented. Additionally, to ensure the model's robustness against a variety of environmental conditions, min-max scaling is performed on the linear acceleration and angular velocity values. This is necessary because the magnification of these values could vary even with a slight twist in the IMU sensor attachment area.

\begin{table}[!t]
    \centering \resizebox{0.8\linewidth}{!}{
    \begin{tabular}{|c|c|c|c|c|c|c|}
    \hline
        ~ & IS & IC & OC & KS & Pooling & OS \\ \hline
         Block 1 & 6×200×1 & 6 & 10 & 5×1 & 2×1 & 10×98×1 \\ \hline
         Block 2 & 10×98×1 & 10 & 20 & 5×1 & 2×1 & 20×47×1 \\ \hline
         Block 3 & 20×47×1 & 20 & 20 & 3×1 & - & 20×45×1 \\ \hline
         Block 4 & 20×45×1 & 20 & 30 & 3×1 & - & 30×43×1 \\ \hline
         Block 5 & 30×43×1 & 30 & 30 & 3×1 & - & 30×41×1 \\ \hline
         Block 6 & 30×41×1 & 30 & 40 & 3×1 & - & 40×39×1 \\ \hline
         Block 7 & 40×39×1 & 40 & 40 & 3×1 & - & 40×37×1 \\ \hline
         Block 8 & 40×37×1 & 40 & 50 & 3×1 & - & 50×35×1 \\ \hline
         Block 9 & 50×35×1 & 50 & 50 & 3×1 & - & 50×33×1 \\ \hline
    \end{tabular}}
    \caption{Feature network structure of the GPR model detailing Input size(IS) and Output size(OS), Input channel(IC) and Output channel(OC), Kernel size(KC), Pooling, and Output size(OS) for each block.}
    \label{table:featnet_gaitphase}
\end{table}
\begin{table}[!t]
    \centering\resizebox{0.8\linewidth}{!}{
    \begin{tabular}{|c|c|c|c|c|c|c|}
    \hline
        ~ & IS & IC & OC & KS & Pooling & OS \\ \hline
         Block 1 & 6×200×1 & 6 & 10 & 5×1 & 2×1 & 10×98×1 \\ \hline
         Block 2 & 10×98×1 & 10 & 20 & 5×1 & 2×1 & 20×47×1 \\ \hline
    \end{tabular}}
    \caption{Feature network structure of the TC model}
    \label{table:featnet_terrain}
\end{table}

\subsection{Model Structure}
In order to design a model capable of handling multiple tasks simultaneously, we construct the model with two main components: the backbone feature network and the head network. The backbone feature network is responsible for reflecting feature information between tasks, while the head network performs a specific task.

The backbone feature network consists of convolutional blocks, which include convolutional kernels, batch normalizations, Rectified linear unit (ReLU) activation functions, and maxpooling layers. Each convolutional block extracts and processes features from the input data, with output dimensions varying depending on specific block configurations, as detailed in Tables \ref{table:featnet_gaitphase} and \ref{table:featnet_terrain}.

The head network is a Multi layer perceptron (MLP) structure comprising fully connected layers and activation functions. Both models use ReLU as the activation function in the hidden layers, while in the output layer, the TC model utilizes softmax, and the GPR model utilizes ReLU. The head network processes the features from the backbone feature network and maps them to the desired output spaces, such as GPR or TC.

Figure \ref{subfig:model_structure} provides an overview of the overall model architecture. The GPR head network connects to the 9th convolutional block of the feature network, while the TC head network links to the 2nd convolutional block of the feature network. As a result, the two models share two convolutional block networks.

\subsection{Labeling Algorithm for GPR Model}
\begin{figure}[!t]
    \centering
    \begin{minipage}{1.0\linewidth}
    \centering
        \begin{subfigure}{1.0\linewidth}
        \centering
        \includegraphics[width=0.9\linewidth]{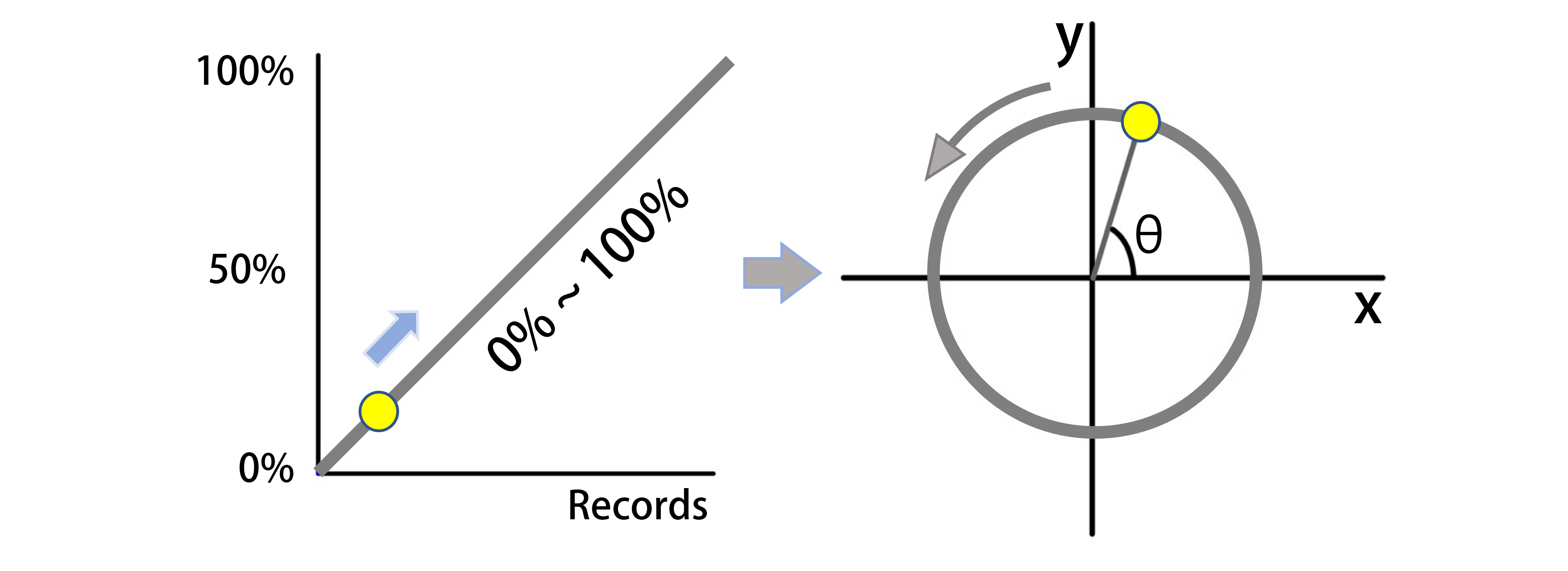}
            \caption{Gait phase output conversion by changing coordinate system: The inherent discontinuity in gait phase output (100\% equals 0\%) is resolved by converting the signal to (x, y) coordinates for gait phase representation. The illustration and concepts were obtained from \cite{phasevariable}.}
            \label{subfig:coordinate_conversion}
        \end{subfigure}
        
        \centering
        \begin{subfigure}{1.0\linewidth}
        \centering
        \includegraphics[width=0.9\linewidth,height=0.3\linewidth]{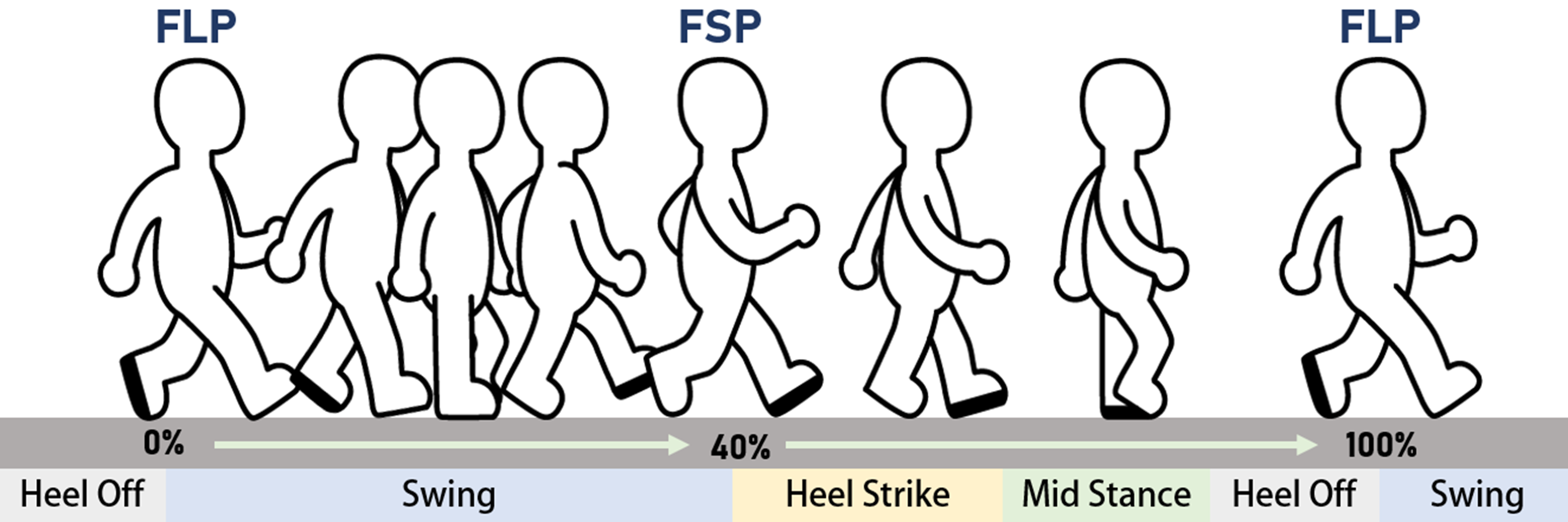}
        \caption{LW terrain gait phase labeling illustration: Using FSR sensors, four discrete phase sections are identified. Based on the left foot, FLP and FSP are assigned 0\% (or 100\%) and 40\%, respectively. Labels are then assigned linearly and iteratively throughout the gait cycle.}
        \label{subfig:fsr_0123}
        \end{subfigure}
        
        \centering
        \begin{subfigure}{1.0\linewidth}
        \includegraphics[width=1.0\linewidth]{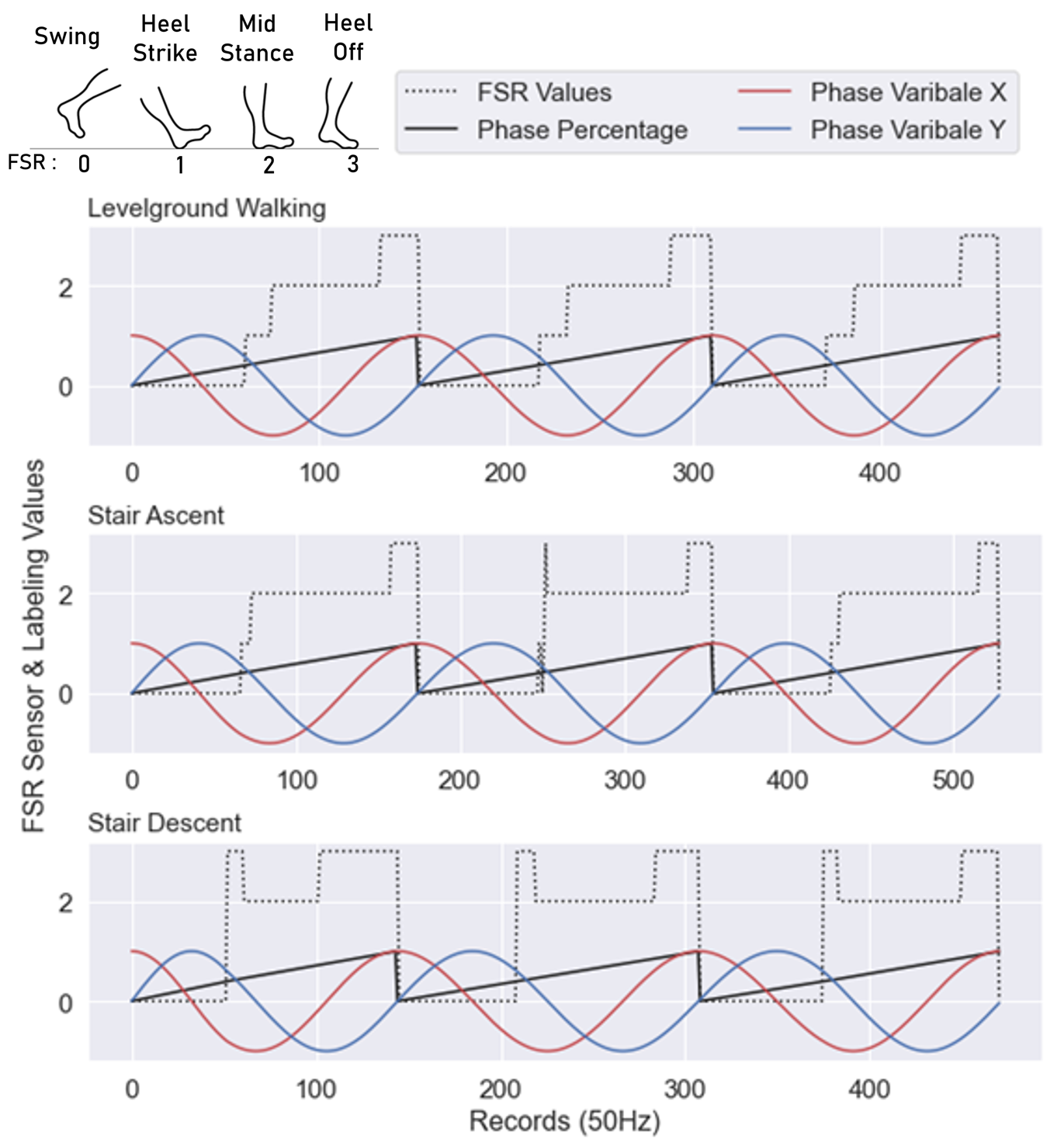}
        \caption{FSR sensor values for three terrain types, along with corresponding phase variable and percentage labeling. Phase events after the swing section vary based on terrain conditions. Original gait percentages range from 0\% to 100\%, but they are scaled down to 0 to 1 for representation alongside FSR values in the graph.}
        \label{fig:labeling_all_img}
        \end{subfigure}
        
    \end{minipage}

    \caption{Illustration of the labeling process based on phase variables derived from FSR sensor values.}
\end{figure}

The TC task does not require an additional labeling process since data collection was carried out separately for each terrain type. However, the GPR task we aim to address in this paper is the continuous GPR task, ranging from 0\% to 100\%, so a well-designed labeling algorithm is required to create a high performance GPR task model. 

We label the gait phase by utilizing an a Force sensitive resistor (FSR) insole sensor worn on the left foot. The FSR insole sensor functions as a foot switch sensor, attaching to both the front and back of the foot, enabling the identification of four distinct gait phase sections: swing, heel strike, mid-stance, and heel off, as illustrated in Figure \ref{subfig:fsr_0123} \ref{fig:labeling_all_img}.

Within the entire gait phase, post-processing is performed to identify the Foot lifting points (FLP) and the Foot stepping points (FSP). In one gait phase cycle, FLP refers to the moment when the foot detaches from the ground for the first time, and FSP refers to the moment when the foot attaches to the ground for the first time. The detected FLP points are fixed at 0\% (or 100\%) in the entire gait percentage section, and the FSP points are fixed at 40\%. The entire gait percentage section is then linearly and iteratively labeled, as illustrated in Figure \ref{fig:labeling_all_img}.

To prevent an increase in bias error due to the discretized point between 100\% and 0\%, the labeled phase values are converted into two continuous phase variables, $x$ and $y$, using the following Equations \ref{eq:phase_conversion_xy}:

\begin{align}
    \theta &= \frac{\text{gait percentage} \times 2\pi}{100}, \nonumber \\
    x &= \cos(\theta), \label{eq:phase_conversion_xy} \\
    y &= \sin(\theta). \nonumber
\end{align}
These continuous phase variables, $x$ and $y$, can then be used to represent the gait phase in a cyclic manner without any discrete point, as described in Figure \ref{subfig:coordinate_conversion}.

To recover the original gait phase percentage from the phase variables x and y, the following Equation \ref{eq:phase_conversion_phase} can be used :

\begin{equation}
\text{gait percentage} = \left(\left(\tan^{-1}\left(\frac{y}{x}\right) + 2\pi\right) \bmod 2\pi\right) \times \frac{100}{2\pi}.
\label{eq:phase_conversion_phase}
\end{equation}
By using this transformation, the GPR task can be learned for the entire gait cycle effectively, as illustrated in Figure \ref{fig:labeling_all_img}, \ref{fig:result_gaitphase}.
The source of the Equation \ref{eq:phase_conversion_xy},\ref{eq:phase_conversion_phase} are from \cite{phasevariable}

\subsection{Training Process}
The training process for implementing the Multitask model consists of two stages: first, training a GPR model, and then training a TC model using the feature network of the GPR model.

Initially, we develop a well-performing GPR model using the designed labeling and input pipelining algorithms. Once the GPR model has achieved satisfactory performance, we connect the input of the TC head network to the output of the GPR model's second convolutional layer. To enable the model to perform both tasks simultaneously, the weights of the backbone network are kept constant, and only the head network is trained.

All the hyperparameters required for training are described in Table \ref{table:hyperparameter}:

\begin{table}[h]
\centering\resizebox{0.8\linewidth}{!}{
\begin{tabular}{  | c || c | c |} 
         \hline
         Task  & GPR &  TC \\ 
         \hline
         Optimizer      & Adam      & Adam \\ \hline
         Learning rate  & 0.0001    & 0.0001 \\ \hline
         loss Function  & MeanSquaredError & CrossEntropy \\ \hline
         Batch Size  & 128 & 128 \\ \hline
         Epoch      &   20   &   10 \\ \hline
        \end{tabular}}
        \caption{Hyperparameters used for training the GPR and TC models, including the optimizer, learning rate, loss function, batch size, and number of epochs.}
        \label{table:hyperparameter}
\end{table}

\section{Experiments and Results}

\subsection{Data Recording Process}
\subsubsection{Subjects}
The experiment involved two healthy male participants. Information about the two subjects is provided in Table \ref{table:subjects}:

\begin{table}[h]
    \centering\resizebox{0.6\linewidth}{!}{
     \begin{tabular}{  | c | c | c | c|} 
     \hline
      & Age(years) & Height(m) & Weight(kg)  \\ 
     \hline
     1 & 26 & 1.71 & 65 \\ \hline
     2 & 34 & 1.73 & 72 \\ \hline
    \end{tabular}}
    \caption{Physical characteristics of the subjects, including age, height, and weight.}
    \label{table:subjects}
\end{table}

\begin{figure}[!t]
    \centering
    \begin{minipage}{1.0\linewidth}
        \centering
        \begin{subfigure}{1.0\linewidth}
        \centering
        \includegraphics[width=0.55\linewidth,height=0.5\linewidth]{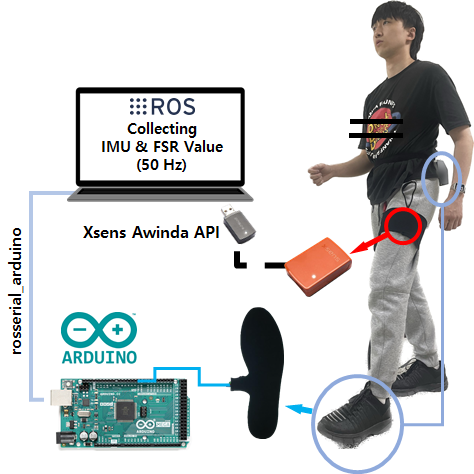}
        \caption{The Xsens mtw awinda dongle connected to the laptop receives sensor values from the Inertial measurement unit(IMU) and transmits them to the Robot operating system(ROS) environment via the API. The insole Force sensitive resistor(FSR) sensor circuit is connected to an arduino mega board, and its sensor values are communicated with the ROS through the rosserial\_arduino package.}
        \label{fig:sensorillustration}
        \end{subfigure}
        \hfill
        \begin{subfigure}{1.0\linewidth}
        \centering
        \includegraphics[width=0.6\linewidth,height=0.5\linewidth]{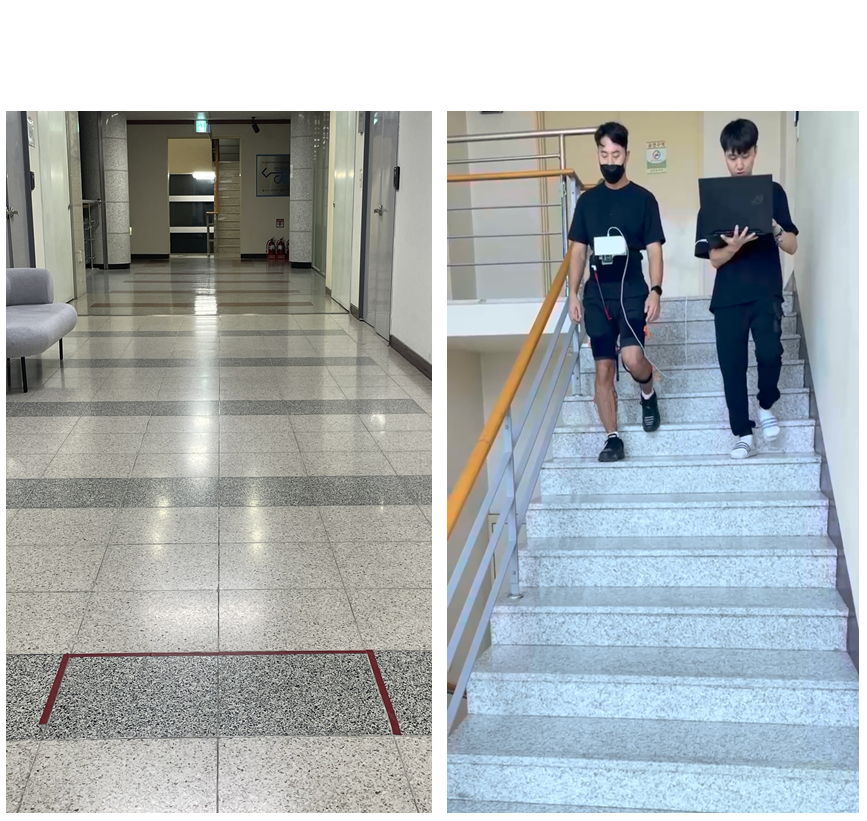}
        \caption{This picture illustrates the terrains used in our experiments:Level-ground walking(LW), Stair ascent(SA), and Stair descent(SD). Data was collected on a 35m long flat surface and a 4.2m long staircase with a 29.05\textdegree angle.}
        \label{subfig:Terrain_img}
        \end{subfigure}
    \end{minipage}
    \caption{Illustration of the data recording process.}
\end{figure}

\subsubsection{Sensors}
We utilized an Inertial measurement unit (IMU) (47×30×13mm, Xsens Mtw Awinda) and FSR insole sensor (90×270×5.3mm, Hexar humancare). The IMU was attached to the left thigh, approximately 17cm above the knee, and the FSR was attached to the left foot, Both sensors data were synchronized at 50Hz using the Robot operating system(ROS, Ubuntu 20.04, Noetic) environment, as described in Figure \ref{fig:sensorillustration}.

\subsubsection{Recording Protocol}
Data was collected under four different speed conditions and three different terrain conditions. The speed conditions were 70, 90, 110, and 130 Beats per minute (BPM), while the terrain conditions included Level-ground walking (LW), Stair ascent (SA), and Stair descent (SD). Data was gathered on a 35-meter-long flat surface and a 4.2-meter-long staircase with a 29.05\textdegree angle, as illustrated in Figure \ref{subfig:Terrain_img}.

\subsection{Training \& Test Dataset Setting}
The training and test dataset settings differed for the GPR and TC task models. For the GPR model, which requires the model to form a rich feature representation network, we randomly split the train and test datasets in a 9:1 ratio.

On the other hand, in the case of TC, we conducted the experiment under extremely limited training data conditions to effectively validate the model's performance. This approach allowed us to assess data efficiency in scenarios resembling real-world situations where acquiring large amounts of data may not be practical. The training dataset for TC was created by randomly selecting five step cycles from each terrain, totaling only 15 step cycles. The test dataset consisted of the remaining data not included in the 15 selected steps.

Both the TC and GPR models used five random seed values to split the training and test sets randomly.

\subsection{Data Efficiency Validation}
The primary goal of this paper is to verify whether the knowledge obtained from a well-trained GPR model can facilitate learning for the new TC task. To effectively demonstrate this data efficiency, we evaluated the performance of the TC model by comparing it with three different model cases:
\begin{itemize}
\item Model 1: A model trained for the TC task using a pretrained feature network from the GPR.
\item Model 2: A model with the same structure as Model 1 but trained from scratch without a pretrained network.
\item Model 3: A head network model, MLP.
\end{itemize}
By comparing the performance of Model 1, which utilizes the GPR feature network, with the performance of the two other baseline models that do not, we validated the data efficiency of our Multitask learning approach.

\begin{figure*}[!t]
    \centering
    \begin{minipage}{0.43\linewidth}
        \centering
        \begin{subfigure}{1.03\linewidth}
        \centering
        \includegraphics[width=1.0\linewidth,height=0.6\linewidth]{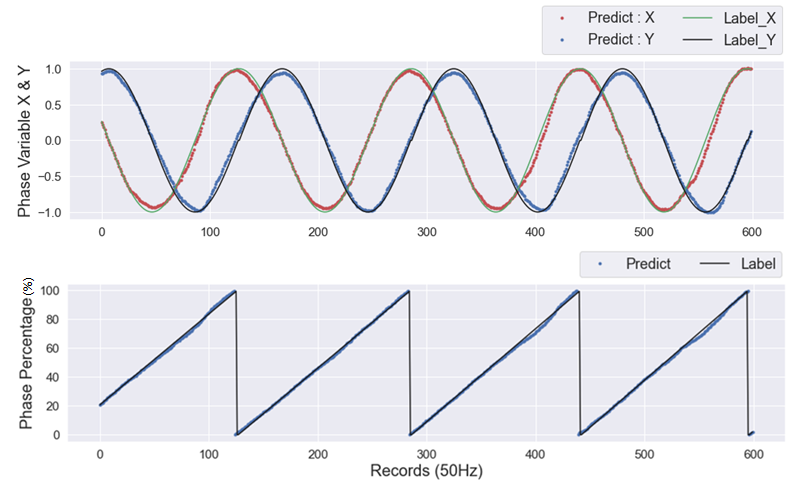}
        \caption{Output results of the well-trained GPR model. gait percentage values are obtained by applying Equation \ref{eq:phase_conversion_phase} to phase variable $x$ and $y$ values.}
        \label{fig:result_gaitphase}
        \end{subfigure}
        \begin{subfigure}{1.0\linewidth}
        \centering
        \includegraphics[width=1.1\linewidth,height=0.42\linewidth]{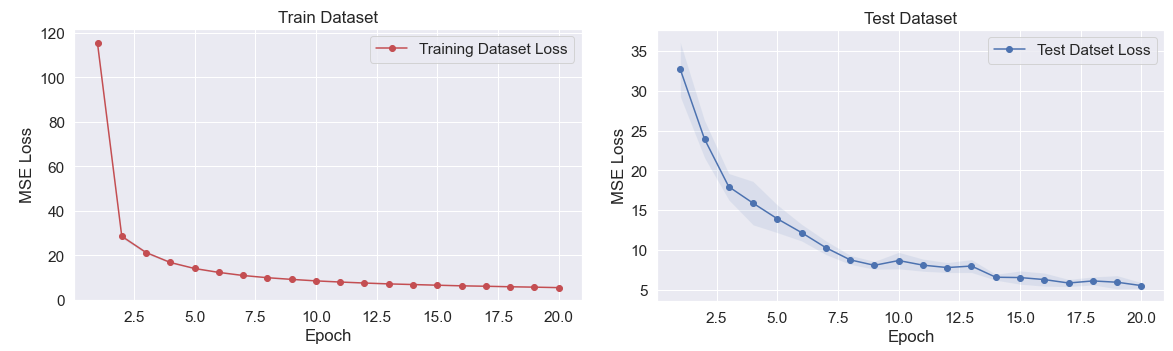}
        \caption{This figure demonstrates the MSE loss values for gait phase training and test dataset, with the trained gait phase recognition model achieving an RMSE below 3\%, indicating well-performing model.The detailed loss values for GPR model can be found in Figure \ref{tab:total_performance}.}
        \label{fig:result_gaitphase_loss}
        \end{subfigure}
    \end{minipage}
    \hspace{0.05\linewidth}
    \begin{minipage}{0.5\linewidth}
        \centering
        
        \begin{subfigure}{1.0\linewidth}
        \centering
        \includegraphics[width=1.0\linewidth,height=0.40\linewidth]{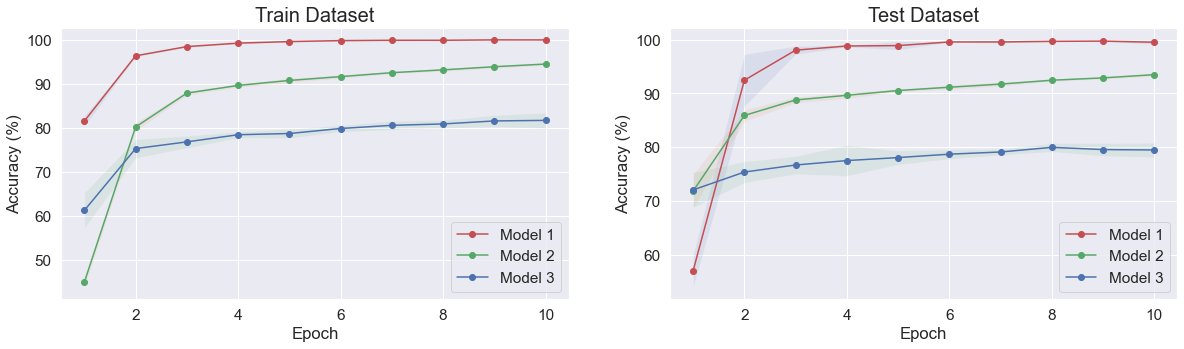}
        \caption{This figure illustrates the accuracy performance of the three model cases for Terrain classification. It can be seen that our proposed Model 1 achieved 99.5 \(\pm\) 0.044\% accuracy, outperforming the other two baseline models with performance of 93.5 \(\pm\) 0.031\% and 79.5 \(\pm\) 0.013\%, respectively. The detailed accuracy performance values for each model can be found in Figure \ref{tab:total_performance}. }
        \label{fig:result_terrain}
        \end{subfigure}
        \hfill
        \begin{subfigure}{1.0\linewidth}
        \centering
        \includegraphics[width=1.0\linewidth,height=0.40\linewidth]{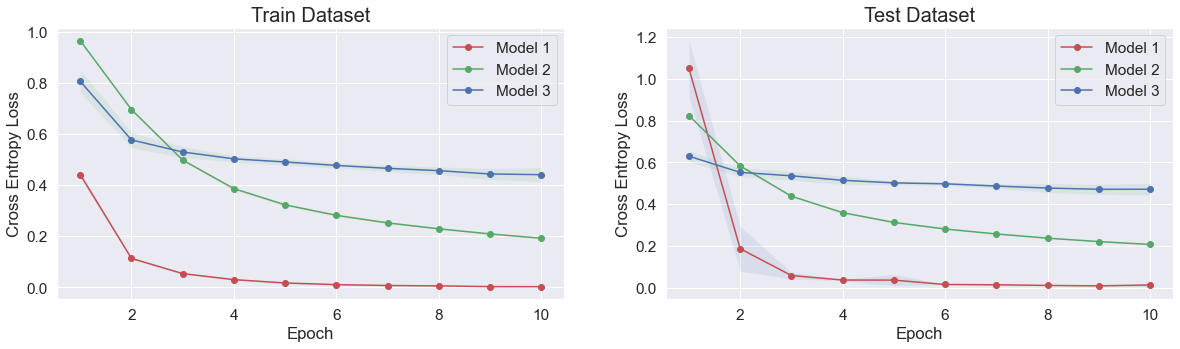}
        \caption{This figure compares the cross entropy loss performance of the three model cases, with model 1 demonstrating the fastest reduction in loss value. The detailed loss values for each model can be found in Figure \ref{tab:total_performance}.}
        \label{fig:result_terrain_loss}
        \end{subfigure}
    \end{minipage}
    
    \caption{An overview of the training results and performance of our proposed Multitask learning approach. Figure \ref{fig:result_gaitphase} illustrates the actual output visualization of the GPR model, while Figure \ref{fig:result_gaitphase_loss} presents the average MSE loss values for the GPR model on the training and test datasets. The comparison of accuracy performance and cross-entropy loss performance for the three TC model cases is depicted in Figures \ref{fig:result_terrain} and \ref{fig:result_terrain_loss}.}
    \label{fig:overall_result}
\end{figure*}

\begin{figure}[!t]
\centering
\includegraphics[width=1.0\linewidth]{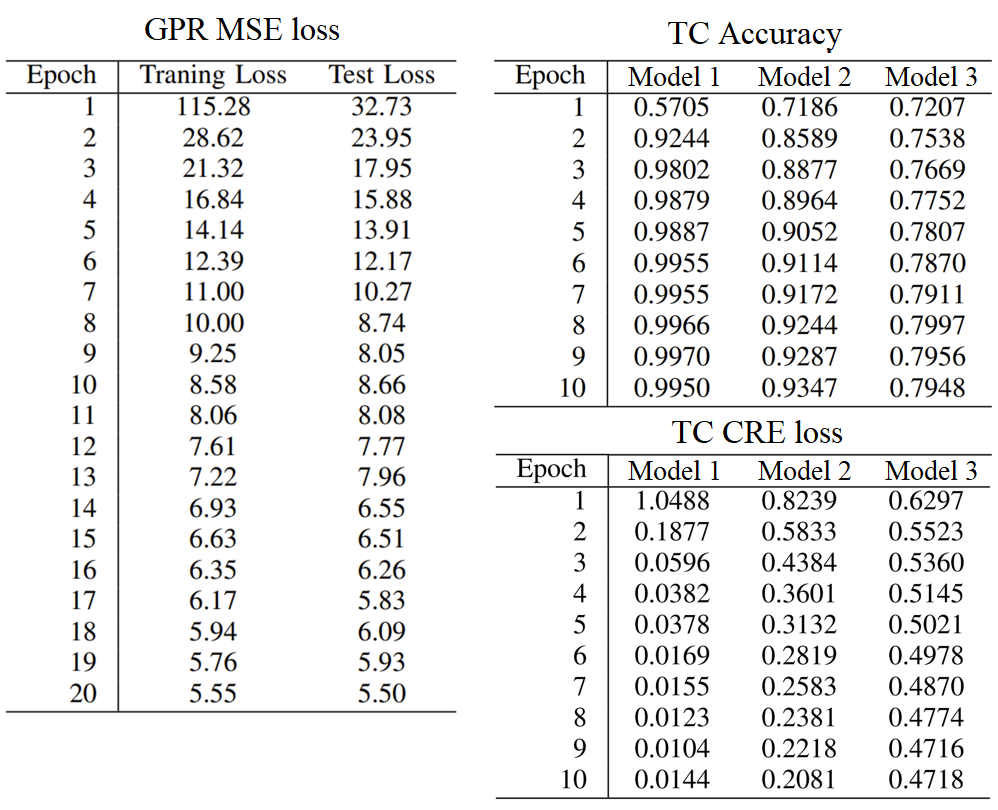}
\caption{Performance summary of TC \& GPR models in a table format. For GPR, the overall model learning is demonstrated to be well-performed through the MSE loss value. In the case of TC, the accuracy and cross-entropy loss values are used to compare the performances of the three models. Model 1, which shares the feature network between tasks, exhibits the best performance.}
\label{tab:total_performance}
\end{figure}

\subsection{Results}
The entire learning results are illustrated in Figure \ref{fig:overall_result} and detailed in Figure \ref{tab:total_performance}. The GPR model achieved an average RMSE value of 2.345 \(\pm\) 0.08\% on the final epoch, indicating well-performing recognition performance.

The TC model, which utilized the feature networks of the GPR model, achieved a final accuracy of 99.5 \(\pm\) 0.044\%, outperforming the other two baseline models' performance 93.5 \(\pm\) 0.031\%, 79.5 \(\pm\) 0.013\%. Under limited dataset conditions, the comparison results demonstrate the potential for data efficiency in our Multitask learning approach, as the TC model utilizing the feature network of the GPR model surpassed the performance of the other two baseline models.
These results suggest that by utilizing a knowledge-sharing backbone feature network for solving multiple tasks, the model can achieve improved data efficiency compared to relying on independent models for each recognition task in the context of lower-limb exoskeleton robots.

The successful recognition performance of the CNN model in human walking movement data indicated that appropriate feature extraction was being carried out through functional operations of the convolutional kernel within the data of human movement \cite{cnn_survey}. The success of our proposed Multitask learning approach can be interpreted in a similar context. It suggests that the feature extraction within the convolutional layers for both tasks may involve similar functional operations related to human gait and environmental interactions. 
% Future research can explore these functional operations in greater detail, enabling more fundamental advancements in Multitask learning.

Although our results demonstrate the effectiveness of our approach for addressing recognition challenges in lower-limb exoskeleton robots, its applicability is limited by the simplicity of the TC task. To further validate our approach, future studies should involve a wider variety of tasks that better represent real-world scenarios. The Multitask learning approach could be extended to broader recognition problems, such as Human activity recognition(HAR) \cite{Human_Activity_Recognition}. 
% By examining the utility of multitask learning approaches in more complex tasks, future research can investigate the generalizability of our framework across different tasks and enhance the adaptability of lower-limb exoskeleton robots in real-world settings.

\section{Conclusions}
In this study, we first designed a suitable input pipelining algorithm for CNN models and converted the input data into a three-dimensional format in the form of an image with six channels, which is expected to work well with convolutional kernels (Figure \ref{subfig:input_pipeline}).
To handle multiple tasks simultaneously, we designed and trained a CNN model to have sufficient feature representation power and enable sharing of features between the two tasks (Figure \ref{subfig:model_structure}, Table \ref{table:featnet_gaitphase} and \ref{table:featnet_terrain}).
We labeled the gait phase values linearly between 0-100\% using FSR sensors (Figure \ref{fig:labeling_all_img}) and utilized Equation \ref{eq:phase_conversion_xy},\ref{eq:phase_conversion_phase} to prevent bias error caused by discrete points between 100\% and 0\%.
The training process was carried out with the hyperparameter settings in Table \ref{table:hyperparameter}. we first trained the GPR model and then connected a portion of its feature network to the TC head network to learn the TC task.
To demonstrate the data efficiency of our Multitask learning approach, we compared the accuracy and cross-entropy loss performance index of the proposed TC model to other baseline models, showcasing the potential data efficiency of our Multitask learning approach.
As a results, the GPR model, created using our designed approach, exhibited well-trained performance with an RMSE value of 2.345 \(\pm\) 0.08\%. and the TC model, which utilized the GPR model's feature network, achieved an accuracy of 99.5 \(\pm\) 0.044\% outperforming the other two baseline models. Through this comparison process, we were able to demonstrate the effectiveness of Multitask learning in achieving data efficiency. (Figure \ref{fig:overall_result} and \ref{tab:total_performance})

By examining the utility of Multitask learning approaches in more complex tasks, future research can investigate the generalizability of our framework across different tasks and enhance the adaptability of lower-limb exoskeleton robots in real-world settings.

% \section{Discussion}
% The successful recognition performance of the CNN model in human walking movement data indicates that appropriate feature extraction is being carried out through functional operations of the convolutional kernel within the data of human movement \cite{cnn_survey}. The success of our proposed Multitask learning approach can be viewed in a similar context. It suggests that the feature extraction within the convolutional layers for both tasks may involve similar functional operations related to human gait and environmental interactions. Future research can explore these functional operations in greater detail, enabling more fundamental advancements in Multitask learning.

% Although the results demonstrate the effectiveness of our approach for recognition challenges in lower-limb exoskeleton robots, its applicability is limited by the simplicity of the TC task. To further validate our approach, future studies should involve a wider variety of tasks that better represent real-world scenarios. The Multitask learning approach could be extended to broader recognition problems, such as Human activity recognition(HAR) \cite{Human_Activity_Recognition}. By examining the utility of Multitask learning approaches in more complex tasks, future research can investigate the generalizability of our framework across different tasks and enhance the adaptability of lower-limb exoskeleton robots in real-world settings.

\bibliographystyle{unsrt} % We choose the "plain" reference style
\bibliography{references} % Entries are in the refs.bib file

\end{document}